%% file: main.tex
\documentclass[11pt]{article}

\usepackage[]{acl}

\usepackage{times}
\usepackage{tcolorbox}
\usepackage{amsmath}
\usepackage[utf8]{inputenc} 
\usepackage[T1]{fontenc}   
\usepackage{hyperref}       
\usepackage{url}
\usepackage{multirow}       
\usepackage{booktabs}  
\usepackage{amssymb}
\usepackage{amsthm}

\usepackage{amsfonts}       
\usepackage{nicefrac}       
\usepackage{tabularx}
\usepackage{longtable}
\usepackage{supertabular}
\usepackage{makecell}
\usepackage{microtype}      
\usepackage{xcolor}         
\usepackage{tabularx}
\usepackage{graphicx}
\graphicspath{ {imgs/} }
\usepackage{svg}
\usepackage{enumitem}
\usepackage{subcaption}
\usepackage{amsthm, amsmath}
\usepackage[linesnumbered,ruled]{algorithm2e}
\usepackage{colortbl}
\usepackage{balance}

\definecolor{blue}{RGB}{17,220,247}
\definecolor{purple}{RGB}{163,115,250}
\definecolor{caribbeangreen}{rgb}{0.0, 0.8, 0.6}

\definecolor{GREEN}{RGB}{84,130,53}

\usepackage{wrapfig}

\usepackage{bbm}

\usepackage{algpseudocode}

 \DeclareMathOperator*{\argmin}{arg\,min}

\newcommand{\name}{\textsc{LiveFC}}

\definecolor{GREEN}{RGB}{84,130,53}

\usepackage{bbm}

\usepackage{algpseudocode}

\usepackage{pgfplots}
\pgfplotsset{compat=1.15}
\usepgfplotslibrary{
  statistics,
  colorbrewer,
  groupplots,
}
\usetikzlibrary{
  patterns,
  shapes.geometric,
  decorations.text,
  matrix,
  fit,
  backgrounds,
  positioning,
}
\tikzset{
  fignode/.style={
    outer sep=0.25em,
  }
}
\tikzset{
  framedfignode/.style={
    outer sep=0.25em,
    inner sep=0.5em,
    rounded corners,
    draw,
  }
}
\colorlet{plotColorNeutral}{gray}
\definecolor{plotColor1}{HTML}{f61a1c}
\definecolor{plotColor2}{HTML}{377eb8}
\definecolor{plotColor3}{HTML}{4daf4a}
\definecolor{plotColor4}{HTML}{984ea3}
\colorlet{plotColorNeutral*}{plotColorNeutral!40}
\colorlet{plotColor1*}{plotColor1!60}
\colorlet{plotColor2*}{plotColor2!60}
\colorlet{plotColor3*}{plotColor3!60}
\colorlet{plotColor4*}{plotColor4!60}
\pgfplotsset{
    colormap={greenred}{HTML=(4daf4a) HTML=(e41a1c)},
    colormap={redgreen}{HTML=(e41a1c) HTML=(4daf4a)}
}

\theoremstyle{definition}

\title{LiveFC: A System for Live Fact-Checking of Audio Streams}

\author{ Venktesh V \\
  Delft University of Technology \\
  Delft, Netherlands \\
  \texttt{v.viswanathan-1@tudelft.nl} \\\And
 Vinay Setty \\
  University of Stavanger,   Factiverse AI \\
 Stavanger, Norway \\
  \texttt{vsetty@acm.org} \\}

\begin{document}
\maketitle
\begin{abstract}

\input{abstract.tex}

\end{abstract}

\input{intro-AA}

\input{03-system_design}

\input{demonstration}

\input{User_interface}
\input{related-work}
\input{conclusion}
\newpage
\clearpage 

\bibliography{references}
\bibliographystyle{acl_natbib}

\input{appendix}

\end{document}

%% file: abstract.tex
The advances in the digital era have led to rapid dissemination of information. This has also aggravated the spread of misinformation and disinformation. This has potentially serious consequences, such as civil unrest. While fact-checking aims to combat this, manual fact-checking is cumbersome and not scalable. While automated fact-checking approaches exist, they do not operate in real-time and do not always account for spread of misinformation through different modalities. This is particularly important as proactive fact-checking on live streams in real-time can help people be informed of false narratives and prevent catastrophic consequences that may cause civil unrest. This is particularly relevant with the rapid dissemination of information through video on social media platforms or other streams like political rallies and debates. Hence, in this work we develop a platform named \name{}, that can aid in fact-checking live audio streams in real-time. \name{} has a user-friendly interface that displays the claims detected along with their veracity and evidence for live streams with associated speakers for claims from respective segments. The app can be accessed at \url{http://livefc.factiverse.ai} and a screen recording of the demo can be found at \url{https://bit.ly/3WVAoIw}.


%% file: intro-AA.tex
\section{Introduction}
\label{sec:intro}

The rapid proliferation of misinformation and disinformation in the digital era has lasting impacts on society, politics, and the shaping of public opinion. While several efforts have been undertaken to combat misinformation with the support of manual fact-checkers in platforms such as Politifact, it is not scalable at the current rate of growth in misinformation. Hence, automated fact-checking approaches have been proposed \cite{guo-etal-2022-survey,trustworthy} which has made tremendous advances in recent times with advent of deep learning based approaches. Majority of the existing automated fact-checking approaches are primarily focused on textual modality \cite{FCsurvey,guo-etal-2022-survey,political_debates}. However, real-world misinformation and disinformation can be spread through multiple possible modalities, such as audio, video, and images \cite{Mocheng,akhtar-etal-2023-multimodal}. It has also been observed that multi-modal content has higher engagement and spreads faster than text only content \cite{multimodal_spreads_faster}. Hence, it is crucial to fact-check multi-modal content.

\begin{figure*}
    \centering
    \includegraphics[width=0.95\linewidth]{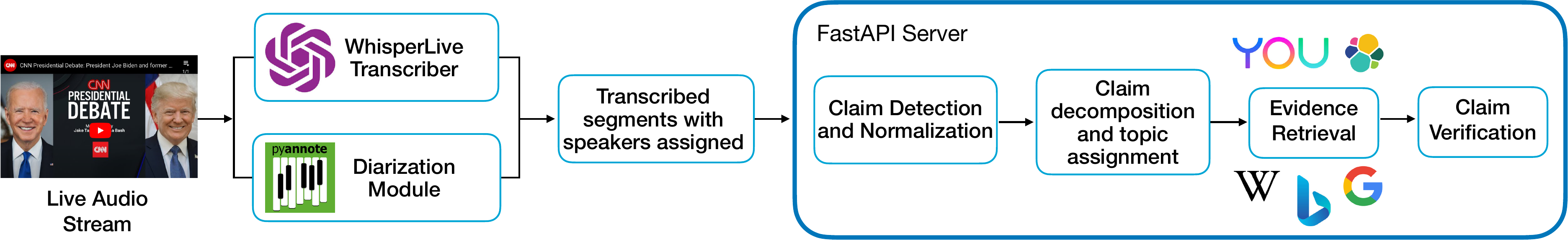}
    \caption{\name{} pipeline for fact-checking live audio streams like political debates.}
    \label{fig:livefc_pipeline}
\end{figure*}

Misinformation spread through multi-modal content, such as political debates, interviews, and election campaigns, is time-critical due to its potential to sway public opinion and its perceived reliability \cite{multimodal_credibility}. Manual fact-checking is cumbersome and time-consuming, and existing automated tools focus on post-hoc verification, which is ineffective against rapidly spreading misinformation. To address this, we developed \name{}, a tool that transcribes, diarizes speakers, and fact-checks spoken content in live audio streams in real-time (within seconds), targeting misinformation at its source. While focused on live events like election debates and campaign rallies, \name{} also works with long-form offline content such as parliament discussions, interviews, and podcasts. Fact-checkers and news reporters find \name{} particularly useful for detecting and verifying claims in real-time. This was validated in a pilot study with Danish fact-checkers Tjekdet\footnote{\url{https://tjekdet.dk}} during the European Parliament election in June 2024, where the tool helped catch important claims that would otherwise have been missed.\footnote{\url{https://factiverse.ai/live}} Additionally, we conducted a case study of the first US presidential debate of 2024, comparing manual fact-checks from the Politifact with those done by \name{}.

Recent advances in automatic speech recognition (ASR) models, like Whisper by OpenAI~\cite{radford2023robust}, have significantly improved audio transcription quality. Existing solutions like Pyannote for speaker diarization \cite{bredin2020pyannote} also perform well, but mainly for offline content. Real-time transcription and speaker diarization for live content pose unique challenges. To enable live fact-checking, we need to transcribe and identify speakers in smaller segments of the audio stream and align speakers with the transcribed text. This demo system showcases techniques to extend Whisper and Pyannote for live-streaming applications.

\name{} architecture is depicted in Figure \ref{fig:livefc_pipeline} which has 6 key components: 1) A transcriber module that can operate on streaming data to transform live audio streams to text, 2) A diarization module that identifies the speaker for the audio segments. 3) A claim detection and normalization module that identifies check worthy claims from the transcribed segments in real-time 4) A claim decomposition and topic assignment module that aids in decomposing claims to questions for reasoning which renders the fact-checking process explainable and assigns a broad set of topics for analysis of the fact-checks 5) An evidence retrieval module that retrieves up-to-date evidence from the web search and past fact-checks and a 6) claim verification component that employs state-of-the-art fine-tuned Natural Language Inference (NLIs) models.

Our key contribution is that the entire pipeline operates in a real-time manner using efficient and effective models, fact-checking claims as they are stated. We posit that this would aid fact-checkers in curbing the spread of misinformation at the source without delay.

\vspace{-10pt}


%% file: 03-system_design.tex
\section{System Design}
An overview of our live fact-checking pipeline \name{} is shown in Figure \ref{fig:livefc_pipeline}. The live audio stream is provided as input to a speaker diarization module and transcription module in parallel. This is followed by a mapping phase where the transcribed segments are mapped to respective speakers based on timestamps and other meta-data. The resulting transcribed segments are then sent to a claim identification module followed by claim decomposition, evidence retrieval and claim verification stages. The pipeline is hosted using a Python FastAPI
backend. The frontend is implemented using the Streamlit framework.\footnote{\url{https://streamlit.io}} Then the claims are verified using evidence retrieval and claim verification components which employ state-of-the-art fine-tuned models. There are several ML models used
in the backend dedicated for (a) check-worthy claim detection, (b) topic categorization, (c) evidence ranking, (d) transcription, (e) speaker diarization and (g) veracity prediction. In addition, we use a self-hosted open large language model (Mistral-7b) with chain of though (CoT) prompting for claim normalization and claim decomposition. These models are also quantized yet effective, enabling real-time processing with low computational requirements.

\subsection{Transcription of Live Audio Stream}

We adapt the Whisper Live\footnote{\url{https://github.com/collabora/WhisperLive}} implementation for our fact-checking pipeline. We use the whisper-large-v3 model, a sequence-to-sequence model pre-trained on a large amount of weakly supervised $(audio, transcript)$ pairs, which directly produces raw transcripts. We process the audio stream in segments to support HLS (HTTP Live Streaming), which is then buffered and transmitted to the transcription client via the FFmpeg encoder.\footnote{\url{https://www.ffmpeg.org}} Unlike traditional systems, Whisper Live employs Voice Activity Detection (VAD) to send data to Whisper only when speech is detected, making the process more efficient and producing high-quality transcripts.

\subsection{Online Diarization Module}
For attribution and offline analysis, linking claims to the corresponding speaker is essential. Our diarization module performs real-time speaker identification, known as online speaker diarization with limited context. \name{} employs an overlap-aware online diarization approach \cite{bredin2021endtoendspeakersegmentationoverlapaware}, involving speaker segmentation and clustering. We adapt the diart module\footnote{\url{https://github.com/juanmc2005/diart}} \cite{diart} for our use, utilizing websockets to stream audio content.

The audio stream is sent via websocket to the diarization server, where it undergoes speaker segmentation using a neural network. Every 500ms, the server processes a 5-second rolling audio buffer and outputs speaker active probabilities $A= {s_1...s_n}$, where $n$ is the number of frames. Speakers with an active probability above a tunable threshold $\tau_{active}$ are identified, while inactive speakers are discarded. This approach effectively handles overlapping speakers, making it ideal for live fact-checking of debates. We set $\tau_{active}=0.65$ to reduce false positives.

The segmentation model's permutation invariance means a speaker may not be consistently assigned the same speaker ID over time. To address this, we use incremental clustering to track speakers throughout the audio stream. Initially, speaker embeddings are created after segmentation for the first buffer, forming a centroid matrix $C$. As the rolling buffer updates, local speaker embeddings $(se_1..se_l)$ are compared to the centroids to assign them using an optimal mapping ($m*$):
\[m* = \argmin_{m\in M} \sum_{i=1}^l d(m(i),se_i)\] Where $M$ is the set of mapping functions between local speakers and centroids, with the constraint that two local speakers cannot be assigned to the same centroid. If the distance between a local speaker embedding and all centroids exceeds a threshold $\Delta_{new}$, a new centroid is created. We set $\Delta_{new}=0.75$ to balance sensitivity, avoiding the misclassification of slight tone changes as new speakers, while ensuring new speakers are accurately identified. 

Speaker IDs are mapped to transcript segments using timestamps from the diarization and transcription components, which are run in parallel for efficiency. We use \textit{pyannote/embedding} computing embeddings and the \textit{pyannote/segmentation-3.0} model for segmentation.

\subsection{Check-Worthy Claim Detection Module}
\begin{table}[htb!!]
\begin{tcolorbox}[title=Prompt: Claim Normalization]
\small
\textbf{Instruction}: Given text in the \{lang\} language, you need to rephrase it in a more formal self-contained way to make it easier for fact-checkers. Remove redundant text, resolve any references to pronouns, dates, and other entities. 
The final generated text must be in the {lang} language with self-contained text with no comments and no other text. Keep original quotes made by someone as much as possible. Remember, this will be used as a input for downstream NLP tasks. 

\textbf{Text}: \{text\}
\end{tcolorbox}
\captionof{figure}{Prompt for claim normalization}
\label{fig:normalization_prompt}
\end{table}

The function of this component is to identify claims from transcribed segments that warrant verification. This entails different sub-tasks as follows.

\noindent \textbf{Sentence segmentation and Claim Normalization}:
We first segment the transcription text into sentences using the Spacy library due to its speed and accuracy. Since speech segments may contain implicit references, we transform the sentences to make them self-contained by resolving co-references and removing any unwanted text from the spoken content. This is performed through a generative LLM (Mistral-7b) using the prompt shown in Figure \ref{fig:normalization_prompt}. We term this step as \textit{claim normalization,} as it yields self-contained candidate claims.
\begin{table}[t!!]
    \small
    \centering
    \begin{tabular}{l|rrrrr}
      \textbf{Split}   & \textbf{NC} & \textbf{C} & \textbf{True} & \textbf{False} &\textbf{Total} \\
      \midrule
       Train  & 609 & 548 & 332 & 196 & 1,076  \\
       Dev  & 38 & 25 & 15 & 10 & 63  \\
       Test  & 62 & 38 & 26 & 12 & 100   

    \end{tabular}
        \caption{Dataset distribution for check-worthy claim detection. NC - Not Check-worthy, C - Checkworthy}
    \label{tab:dataset_1}
\end{table}
The self-contained candidate claims are then passed through a check-worthy claim detection model.  We fine-tune a XLM-RoBERTa-Large model
\cite{conneau-etal-2020-unsupervised}, using datasets
from ClaimBuster \cite{claimbuster} and CLEF CheckThat Lab! \cite{alam-etal-2021-fighting-covid} along with a
dataset collected from Factiverse production system (see Table \ref{tab:dataset_1})
to classify sentences into ‘Check-worthy’ and ‘Not check-worthy’. In addition, we also assign a broad set of topics to the claims for further analysis using an LLM. The prompt for the topic classification is shown in Appendix \ref{appendix:topics}.

\subsection{Claim Decomposition and Evidence Retrieval}
\begin{table}[htb!!]
\begin{tcolorbox}[title=Prompt: Claim Decomposition]
\small
\textbf{Instruction}: Transform the given claim into questions to verify the \
veracity of the given claim and find the potential correct facts from search engines.\
The questions must be cover all aspects of the claim and must be generated only in the {lang} language with no translation in the brackets. \
Generate exactly {num\_questions} questions for the claim and prefix the \
questions with \"Question number:\" without making any references to the claim.\
Here are some \textbf{examples}:

\textbf{Examples}:

Claim: "Kelvin Hopins was suspended from the Labor Party due to his membership in the Conservative Party."
\\
Question 1: Was Kelvin Hopins suspended from Labor Party?

Question 2: Why was Kelvin Hopins suspended from Labor Party?

\dots \dots \\
\textbf{Claim}: \{claim\}
\end{tcolorbox}

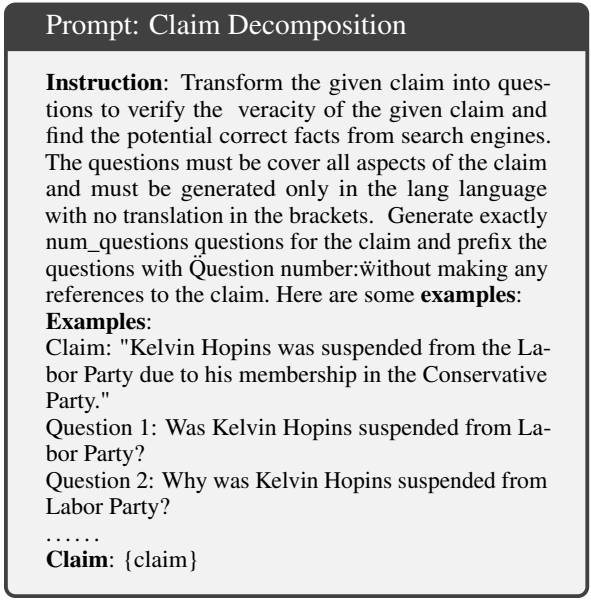
\captionof{figure}{Prompt for claim normalization}
\label{fig:decomposition_prompt}
\end{table}

The main goal is this component is to retrieve high quality evidence for verifying the check-worthy claims from the previous step. Fact-checking is not a linear process and involves multi-step reasoning, where fact-checkers synthesize diverse queries and search the web and other knowledge sources to gather multiple perspectives and evidence to verify a claim. To emulate the process of fact-checkers, we employ a claim decomposition module where we prompt a LLM (Mistral-7b) with few-shot examples to decompose a claim. The prompt is as shown in Figure \ref{fig:decomposition_prompt}.

Following the decomposition step, we retrieve evidence from diverse sources such as Google, Bing,
Wikipedia, You.com, Semantic Scholar8
(contains 212M scholarly articles). Since some claims might be duplicates or similar to existing fact-checked claims, we also search our ElasticSearch index, which houses Factiverse’s fact-checking collection named \textbf{FactiSearch}, which comprises 280K
fact-checks updated in real-time to retrieve related evidence. We filter out evidence from fact-checking sites and deduplicate evidence using meta-data like url, titles and approximate matching of content. We then employ a multilingual cross-encoder model \cite{reimers2019sentencebertsentenceembeddingsusing} (\textit{nreimers/mmarco-mMiniLMv2-L12-H384-v1}) from huggingface to rank the retrieved evidences.
\vspace{-1em}
\subsection{Claim Verification}
Using the ranked evidences, we perform claim verification by formulating the task as a Natural language Inference (NLI) problem. The NLI task involves categorizing whether a claim is supported, refuted by a given piece of evidence or evidence is unrelated to the claim \cite{bowman-etal-2015-large}. We cast this problem to a binary classification task of predicting supported or refuted as we filter out unrelated evidence in the ranking step. We fine-tune an \textit{XLM-Roberta-Large} model from Huggingface on combined data from FEVER \cite{thorne-etal-2018-fever}, MNLI \cite{williams-etal-2018-broad}, X-fact \cite{gupta-srikumar-2021-x} and our collection of real-world fact-checks in \textbf{FactiSearch}. Since each claim has multiple relevant evidence snippets, the NLI model is applied to claim and evidence in a pairwise manner followed by a majority voting phase following prior works \cite{majortiy_vote_1,schlichtkrull2023averitec} to obtain the final verdict. We also summarize the evidence snippets providing justification for the verdict to the user to foster trust in the system. 

%% file: demonstration.tex
\begin{figure*}
    \centering
    \includegraphics[width=1\linewidth]{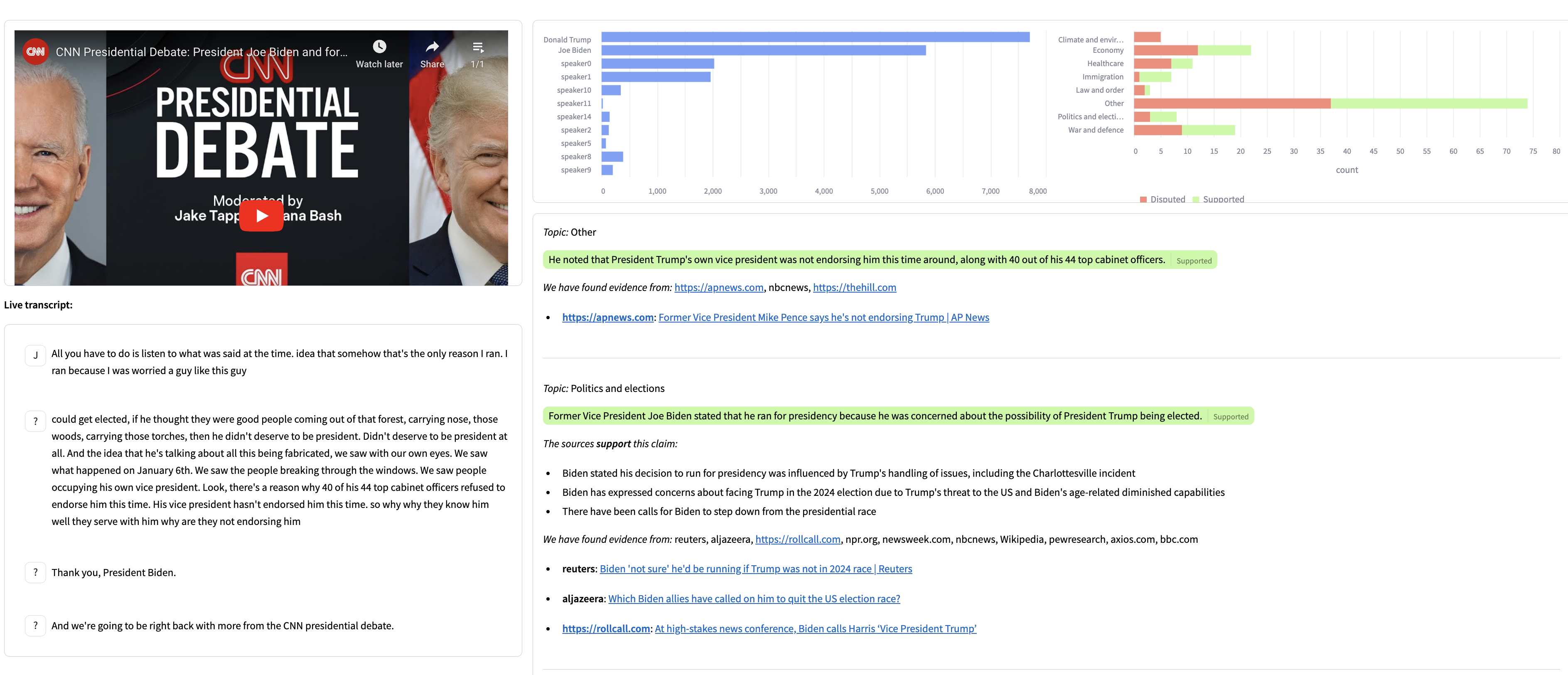}
    \caption{A screenshot of \name{} user interface. The pipeline runs on live stream and the claims detected, their veracity with corresponding evidence appear in real-time.}
    \label{fig:enter-label}
\end{figure*}

\begin{table}

    \centering
    \small
    \begin{tabular}{l|rr|rr}
    \hline
      \textbf{Model}   & \multicolumn{2}{c}{\textbf{Claim Detection}}  & \multicolumn{2}{c}{\textbf{Veracity Prediction}}  \\
         & \textbf{Ma.-F1} & \textbf{Mi.-F1 }& \textbf{Ma.-F1} & \textbf{Mi.-F1 } \\
      \midrule
      Mistral-7b  & 0.590 & 0.600 & 0.526 & 0.527  \\
      
       GPT-3.5-Turbo  & 0.607 & 0.625 &  0.605 & 0.605  \\
      GPT-4  & 0.695 & 0.701 &  0.630 & 0.632  \\ 
       \textbf{Ours}  & \textbf{0.899} & \textbf{0.900} & \textbf{0.708} & \textbf{0.737}   \\
       \hline
    \end{tabular}
        \caption{Claim detection and verification results presented as Micro and Macro-F1 scores for English data.}
    \label{tab:result_summary}
\end{table}

\section{Performance Evaluation}
In this section, we perform offline evaluation of critical components of \name{} using our benchmark for claim detection and verification. We also perform qualitative evaluation of a sample of fact-checks from 2024 presidential debate.
\subsection{Offline Evaluation of Claim Detection and  Verification Components}
For offline evaluation of individual components of \name{} pipeline, we employ the dataset collected from production environment of Factiverse. The statistics of the dataset are shown in Table 1. The prompt employed for the LLM baselines can be found in the Appendix \ref{sec:prompts}. We observe that our fine-tuned XLM-Roberta model outperforms LLM based approaches for tasks of claim detection and verification. We primarily observe that in claim verification, LLMs underperform when compared to smaller fine-tuned models due to their inability to reason and extract required information from evidence and due to hallucination. Hence, we employ our fine-tuned model as part of the pipeline in the \name{} tool.
\input{table-stats}
\input{manual_eval}

\subsection{End to End Evaluation on Live Stream}
We also  evaluate \name{} on the first presidential debate of 2024. A screenshot of the tool is shown in Figure \ref{fig:enter-label}.

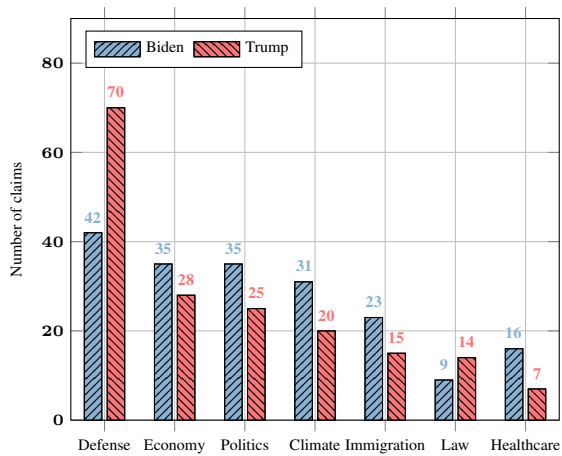
\begin{figure}[!hbt]
    \begin{subfigure}{.8\linewidth}
    \input{topic_stats}
    \end{subfigure}
\caption{Statistics of fact-checks from 2024 presidential debate}
\label{fig:statistics}
\end{figure}

\textbf{Debate Statistics}: We report the statistics obtained from live fact-checking of the debate through our tool \name{}. The number of supported and disputed claims made by each speaker is shown in Table \ref{tab:stats} and topicwise distribution of claims are shown in Figure \ref{fig:statistics}. We observe that topic related to \textit{War and Defense} was the most discussed during the debate. The plot shows the distribution of claims across 7 key topics, and the rest of claims that do not fall into any of these topics are categorized as ``Other" and are not shown in the graph. We also observe that there are a significant number of disputed claims made by the speakers, which highlights the significance of live fact-checking. We also display the evidence and summarize the justification for the veracity label, rendering the process more transparent to the end user.

\textbf{Comparison based evaluation of claims identified and veracity prediction to Politifact}: We compare the claims identified and corresponding predicted labels from manual fact-checker Politifact to those identified and verified by our tool \name{} for the 2024 presidential debate. We observed that we were able to identify all the 30 claims identified by Politifact. We were further able to identify more claims not covered by Politifact which highlights the advantages of automated fact-checking. However, we also acknowledge that some of the claims we identify are false positives and may not be significant enough, which is removed by us in post-processing phase.  When comparing the veracity labels with Politifact for the 30 claims, we observe a macro averaged Precision, Recall and F1 scores of \textbf{82.59, 85.78} and \textbf{83.92}  
 respectively and weighted F1 of \textbf{87.26}. This highlights that \name{} can assist fact-checkers in verifying live-streams at scale.

\textbf{Qualitative evaluation of evidence utility and topic assignments}: We perform a qualitative evaluation of fact-checks performed on 2024 US presidential debate live-stream by sampling 20 claims with retrieved evidence, topic assigned and veracity predictions using our tool \name{}. We requested three annotators with background in automated fact-checking to rate the samples on three factors such as evidence usefulness, evidence completeness and topic relevance using the Likert scale (1-5). The average ratings across annotators with inter-annotator agreement are shown in \textbf{Table \ref{tab:question_quality}}.

%% file: table-stats.tex
\begin{table}[]
\small
\begin{tabular}{l|l|l|l}
\bf Speaker   & \bf Supported               & \bf Disputed                & \bf Total   \\\hline\hline
Trump     & 147 & 205 & 352                        \\
Biden     & 169 & 170 & 339                       \\
\hline

\end{tabular}
\caption{Statistics from fact-checks of 2024 debate}
\label{tab:stats}
\end{table}

%% file: manual_eval.tex
\begin{table}[t!]
 \small

\begin{tabular}{lccc}
\hline
\textbf{EC}  ($\alpha_K$)& \textbf{EU} ($\alpha_K$) & \textbf{TR} ($\alpha_K$)\\ 
\hline

\midrule
 3.46$\pm$1.49 (0.76) & 3.60$\pm$1.39 (0.65)& 4.37$\pm$1.12 (0.51) \\

 
\bottomrule
\end{tabular}
\caption{Manual evaluation. EC: Evidence Completeness, EU : Evidence usefulness, TR : Topic Relevance. We use the Likert scale (1-5) and Kippendorff's alpha ($\alpha_K$) for inter-annotator agreement (in brackets).}
\label{tab:question_quality}
\end{table}

%% file: topic_stats.tex
\begin{tikzpicture}
\edef\mylst{"42","35","35","31","23","9","16"}
\edef\explora{"70","28","25","20","15","14","7"}

    \begin{axis}[
            ybar=2pt,
            width=1.3\textwidth,
            bar width=0.25,
            every axis plot/.append style={fill},
            grid=major,
            xtick={1, 2, 3 , 4 ,5, 6 ,7},
            xticklabels={Defense, Economy, Politics, Climate, Immigration, Law, Healthcare},
            xlabel={},
            ylabel style = {font=\tiny},
            xlabel style = {font=\tiny},
        yticklabel style = {font=\boldmath \tiny,xshift=0.25ex},
        xticklabel style ={font=\tiny,yshift=0.15ex},
            ylabel={Number of claims},
            enlarge x limits=0.08,
            ymin=0,
            ymax=90,
            legend style ={font=\tiny},
            area legend,
            nodes near coords=\pgfmathsetmacro{\mystring}{{\mylst}[\coordindex]}\textbf{\mystring},
            nodes near coords style={font=\tiny,align=center,text width=2.3em},
            legend entries={Biden, Trump},
            legend cell align={left},
            legend pos=north west,
            legend columns=-1,
            legend style={/tikz/every even column/.append style={column sep=0.05cm}},
        ]
        \addplot+[
            ybar,
            plotColor2*,
            draw=black,
            postaction={
                    pattern=north east lines
                },
        ] plot coordinates {
                (1,42)
                (2,35)
                (3,35)
                (4,31)
                (5,23)
                (6,9)
                (7,16)

            };
        \addplot+[
            ybar,
            plotColor1*,
            draw=black,
            nodes near coords=\pgfmathsetmacro{\mystring}{{\explora}[\coordindex]}\textbf{\mystring},
    nodes near coords align={vertical},
            postaction={
                    pattern=north west lines
                },
        ] plot coordinates {
                (1,70)
                (2,28)
                (3,25)
                (4,20)
                (5,15)
                (6,14)
                (7,7)

            };

    \end{axis}
\vspace{-1em}

\end{tikzpicture}

%% file: User_interface.tex
\section{User Interface}

A screenshot of \name{} is shown in Figure \ref{fig:enter-label}. The left pane streams the video/audio, with the lower pane showing the transcribed segments in real-time. On the right pane at the top, we show a summary statistics of time segment of each active speaker along with number of claims made by each person as detected by our check-worthy claim detection module along with the veracity of the claims. We also display a plot demonstrating distribution of claims across different topics. on the lower right panel we display a running list of detected claims along with retrieved evidence and predicted verdict. This ensures the fact-checking process is transparent, as the users can trace the verdict to relevant evidence. 

%% file: related-work.tex
\section{Related Work}
\label{sec:rel-work}
Automated fact-checking approaches have made significant strides in identifying misinformation and assisting fact-checkers and journalists to obviate the time-consuming aspects of manual fact-checking \cite{FCsurvey,trustworthy}.
The existing automated fact-checking approaches primarily focus on text modality \cite{FCsurvey,guo-etal-2022-survey}. However, in the real-world misinformation proliferates through multiple modalities such as audio, images, or video \cite{akhtar-etal-2023-multimodal,Mocheng,Spotfake}. It has also been observed that multi-modal misinformation has a propensity to spread faster, which can have disastrous consequences such as civil unrest or health hazards in context of medical misinformation \cite{multimodal_spreads_faster}. For instance, recent studies on misinformation spread through instant audio messages \cite{audio_disinformation_1, audio_disinformation_2, audio_misinformation} on Whatsapp observed that audio messages are considered to be more reliable. Hence, fact-checking multi-modal information is of crucial importance. While the majority of the existing fact-checking systems primarily focus on text \cite{schlichtkrull2023averitec,political_debates,guo-etal-2022-survey}, more recently focus on multi-modal fact-checking has increased, leading to development of new benchmarks and approaches \cite{spotfakeplus,Spotfake,Mocheng,akhtar-etal-2023-multimodal,rangapur2024finfactbenchmarkdatasetmultimodal}. 

There are fact-checking demos in the literature~\cite{Setty:2024:SIGIRa,Botnevik:2020:SIGIR,popat2018credeye,chern2023factool}, but none of them are designed to fact-checking live content. In this work, we build a tool for live fact-checking of political debates, as political claims play a major role in democracy and sometimes may sway public opinion or cause civil unrest. This has been evidenced by prior study that has demonstrated that fact-checking can help the public make an informed evaluation of political events\cite{citation_politics}. Our tool, considers multiple modalities performing fact-checks in real-time, unlike existing works on political debates which primarily focus on post hoc fact-checking and are limited to textual modality relying on clean transcripts \cite{political_debates,gencheva-etal-2017-context,shaar-etal-2022-role}.
\vspace{-10pt}


%% file: conclusion.tex
\section{Conclusion}

This paper presents the \name{} system, an end to end approach for real-time fact-checking which employs efficient, effective and smaller models. We applied it to the live stream of 2024 political debate and observed that it was able to detect and verify facts in real-time. We conducted offline evaluation of different core components of the system using fact-checking benchmarks. We also conducted manual and qualitative evaluation of fact-checks generated from debate and observed that the system was able to detect all claims detected by manual fact-checkers and also retrieve useful evidence for accurate verification of claims. In the future, we plan to further extend \name{} to handle multi-modal evidence sources. While our current pipeline provides support for multiple languages, we plan to further extend the number of languages covered.
\section{Acknowledgements}
This work is in part funded by the Research Council of Norway project EXPLAIN (grant number 337133).
We would like to acknowledge valuable contributions from Factiverse AI team members Erik Martin who helped develop parts of the backend, Tobias Tykvart who spearheaded development of the Frontend, Sowmya AS for her contributions to the UI, Henrik Vatndal who helped in development and validation of the diarization module and rest of Factiverse AI team for contributions to annotating the fact-checks manually (Maria Amelie, Gaute Kokkvol, Sean Jacob, Christina Monets and Mari Holand).
\section*{Limitations}
While our tool \name{} works well on live-streams, our tool requires the m3u8 format and audios in other formats need to be converted to m3u8 format. Identifying and converting from different formats requires engineering of adaptors, which we reserve for future work. However, currently, other formats can be converted to m3u8 format using existing tools. We plan to build adaptors and provide native support in \name{} in the future. Additionally, our claim verification component currently supports categorizing a claim as ``supported" or ``refuted". In future, we also plan to support other fine-grained categories such as conflicting where a claim is partly true/false.

\section*{Ethics and Impact Statement}
Our live fact-checking tool \name{} aims to assist fact-checkers and journalists to combat misinformation at the source. Since we employ deep learning based methods there is possibility of errors in claim veracity prediction. Hence, we try to render the process as transparent as possible by providing evidence sources, snippets and justification summary used for verification of a claim. This helps the users to look at the sources, evidence snippets and make their own judgement of the veracity. We also do not claim that \name{} would replace manual fact-checkers but would reduce their load and augment their abilities, making fact-checking at scale possible.

%% file: appendix.tex
\clearpage
\newpage

\appendix

\section{Experimental Setup and Prompts}
\label{sec:prompts}

\begin{table}[htb!!]
\begin{tcolorbox}[title=Prompt: Claim Detection]
\small
\textbf{Instruction}: Your task is to identify whether a given text in the \{lang\} language os verifiable using a search engine in the context of fact-checking.
\\
\textbf{Function Definition}: Let's define a function named checkworthy(input: str).

\textbf{Return value}:The return value should be a strings, where each string selects from "Yes", "No".
"Yes" means the text is a factual checkworthy statement.
"No" means that the text is not checkworthy, it might be an opinion, a question, or others.

\textbf{Example}:For example, if a user call checkworthy("I think Apple is a good company.")
You should return a string "No" without any other words, 
checkworthy("Apple's CEO is Tim Cook.") should return "Yes" since it is verifiable.
Note that your response will be passed to the python interpreter, SO NO OTHER WORDS!
Always return "Yes" or "No" without any other words.

checkworthy(\{text\}) 

\end{tcolorbox}
\captionof{figure}{Prompt for LLM based claim verification}
\label{fig:detection_prompt}
\end{table}

\subsection{Checkworthy Claim Detection}
\label{appendix:claim_verification}
The check worthy claim detection prompt used for LLM baselines in Table \ref{tab:result_summary} are shown in Figure \ref{fig:detection_prompt}. For fair evaluation, we set temperature to 0.2 to reduce hallucination for all the LLMs.
\begin{table}[htb!!]
\begin{tcolorbox}[title=Prompt: Claim Verification]
\small
\textbf{Instruction}: You are given a claim and an evidence text both in the {lang} language, and you need to decide whether the evidence supports or refutes. Choose from the following two options.
A. The evidence supports the claim. 
B. The evidence refutes the claim.

For example, you are given

Claim: "India has the largest population in the world."

Evidence: "In 2023 India overtook China to become the most populous country." 
You should return A
Pick the correct option either A or B. You must not add any other words.

Claim: \{claim\} \\
Evidence: \{evidence\} 

\end{tcolorbox}
\captionof{figure}{Prompt for LLM based claim verification}
\label{fig:verification_prompt}
\end{table}

\subsection{Claim verification}
\label{appendix:claim_verification}
The claim verification prompts used for LLM baselines in Table \ref{tab:result_summary} are shown in Figure \ref{fig:verification_prompt}. For fair evaluation, we set temperature to 0.2 to reduce hallucination for all the LLMs.

\subsection{Topic Assignment Prompt}
\label{appendix:topics}

\begin{table}[htb!!]
\begin{tcolorbox}[title=Prompt: Topic Assignment]
\small
\textbf{Instruction}:
Given the text, you need to identify the main topic of the text.

Choose one topic from the following options:

A. War and defence (Ukraine, Palestine, conflicts, foreign policy)\\
B. Economy (Taxes, cost of living)\\
C. Healthcare (abortion, parental rights, insurance)\\
D. Law and order (Police force, gun control, crime)\\
E. Immigration \\
F. Climate and environment (Global warming, pollution, de-carbonization)\\
G. Politics and election (political and election issues)\\
H. Other\\

\textbf{Examples:} \\
\textbf{Text:} There are 20 million people getting healthcare through Obamacare.\\
\textbf{Topic:} C\\
\textbf{Text:} In 2021, 2022, California’s lost 750,000 residents to other states due to cost of living.\\
\textbf{Topic:} B\\

\textbf{Text:} We have a 50-year low in the crime rate. In the last 10 years we’ve had a 45\% decline in homelessness. California has had a 45\% increase in homelessness. \\
\textbf{Topic:} D\\

\textbf{Text:} During our Administration in the Recovery Act, I was able, was in charge, able to bring down the cost of renewable energy to cheaper than or as cheap as coal and gas and oil.\\
\textbf{Topic:} F\\

\textbf{Text:} In American cities, we have protestors calling for global Islamic war and demanding that Israel be wiped off the map, or in the words of Congresswoman Tlaib, “From the river to the sea.”\\
\textbf{Topic:} A\\

\textbf{Text:} The Obama administration did fail to deliver immigration reform, which had been a key promise during the administration. It also presided over record deportations as well as family detentions at the border before changing course. \\
\textbf{Topic:} E\\

\textbf{Text:} Christiano Ronaldo is the best football player in the world.\\
\textbf{Topic:} H\\

\textbf{Text:} Electoral college is a disaster for a democracy.\\
\textbf{Topic:} G\\

Answer only A-H. Do not add any other words. If you are not sure, choose H.\\

Text: {text}\\
Topic: \\
""" 
\end{tcolorbox}
\captionof{figure}{Prompt for LLM based topic assignment}
\label{fig:topic}
\end{table}

To assign topics to claims, we use the Mistral (7b) model by providing examples in the prompt. The prompt employed is as shown in Figure \ref{fig:topic}